\documentclass[conference]{IEEEtran}
\usepackage{times}
\usepackage{graphicx}
\usepackage{multirow}
\usepackage{booktabs}
\usepackage{array}
\usepackage{sidecap}

\usepackage[numbers]{natbib}
\usepackage{multicol}
\usepackage[bookmarks=true]{hyperref}
\renewcommand{\thetable}{\arabic{table}}
\renewcommand{\tablename}{Table}
\usepackage{caption}
\begin{document}

\title{Dreaming Falcon: \\Physics-Informed Model-Based Reinforcement Learning for Quadcopters}

\author{\IEEEauthorblockN{Eashan Vytla, Bhavanishankar Kalavakolanu, Andrew Perrault, and Matthew McCrink}
\IEEEauthorblockA{The Ohio State University\\
Columbus, OH, USA\\
Email: vytla.4@osu.edu}
}



%

\maketitle

\begin{abstract}
Current control algorithms for aerial robots struggle with robustness in dynamic environments and adverse conditions. Model-based reinforcement learning (RL) has shown strong potential in handling these challenges while remaining sample-efficient. Additionally, Dreamer has demonstrated that online model-based RL can be achieved using a recurrent world model trained on replay buffer data. However, applying Dreamer to aerial systems has been quite challenging due to its sample inefficiency and poor generalization of dynamics models.

Our work explores a physics-informed approach to world model learning and improves policy performance. The world model treats the quadcopter as a free-body system and predicts the net forces and moments acting on it, which are then passed through a 6-DOF Runge-Kutta integrator (RK4) to predict future state rollouts.

In this paper, we compare this physics-informed method to a standard RNN-based world model. Although both models perform well on the training data, we observed that they fail to generalize to new trajectories, leading to rapid divergence in state rollouts, preventing policy convergence.
\end{abstract}

\IEEEpeerreviewmaketitle

\section{Introduction}
Modern control algorithms for aerial robots often fall short in dynamic or unpredictable environments. Traditional methods tend to rely heavily on hand-crafted models or assumptions that break down in dynamically adapting environments. Model-based reinforcement learning (MBRL) is a promising alternative. It is sample-efficient and effective at learning complex behaviors under dynamic conditions. \citet{hafner_mastering_2024}, in particular, made online MBRL feasible by training a recurrent world model on replayed trajectories. Then, Dreamer was adapted into DayDreamer to perform well for robotics applications in \citet{wu_daydreamer_2022}. Although the RNN world model performed well on the robot demonstrations in DayDreamer, RNNs have been shown to perform poorly for quadcopter dynamics in \citet{mohajerin_multi-step_2018}, motivating further exploration into physics-informed modeling.

\citet{ramesh_physics-informed_2023} used Lagrangian physics to build a more accurate model for reinforcement learning in robotics. Although this method works well in simple gym environments for rigid-body systems, it requires building a detailed dynamics model for each system, including mass matrices and energy functions. This becomes a major bottleneck for aerial robots, where modeling every component accurately is time consuming and difficult to scale. As a result, the approach is difficult to generalize and adapt quickly to new platforms or flight conditions, limiting its usefulness in real-world aerial applications.

Our hypothesis was that the Dreamer-style world model, when augmented with physics-informed structure, would overcome the generalization challenges typically faced in quadcopter modeling. Specifically, we believed that predicting net forces and moments acting on the free body and integrating them through a known 6DOF dynamics model would provide the inductive bias needed to stabilize world-model learning and enable robust policy training.

Our long-term vision is to eliminate the need for tedious manual tuning of PID gains or other stability parameters during quadcopter deployment. Traditionally, each new flight condition---whether it be wind disturbance, payload change, or motor degradation—necessitates parameter re-adjustment through trial and error. By leveraging a MBRL architecture, we aim to build control policies that continuously adapt to environmental variations and system changes in real time. Instead of re-tuning controllers, operators would simply deploy Dreaming Falcon, which would evolve on-the-fly through online interactions, making autonomous aerial systems more robust, resilient, and easy to deploy across varying tasks and platforms.

To test this hypothesis, we developed a physics-informed world model, which predicts the net forces and moments acting on the system rather than directly predicting future states. The forces and moments are passed through a Runge-Kutta (RK4) integrator using known equations of motion. We compared this approach to a standard RNN-based world model trained directly on state transitions. 

Our results show that while both models perform well on replayed trajectories, they fail to generalize to unseen trajectories. This leads to drift in rollout predictions and prevents policy convergence, especially during transitions between different flight regimes, such as hover-to-forward flight. These findings suggest that even physics-informed inductive bias is not sufficient to ensure generalization in high-dimensional, underactuated systems like quadcopters. In the following sections, we detail our methodology, training setup, and experimental results supporting these insights.

\section{Methods}
\begin{SCfigure}
    \centering
    \includegraphics[width=0.4\linewidth]{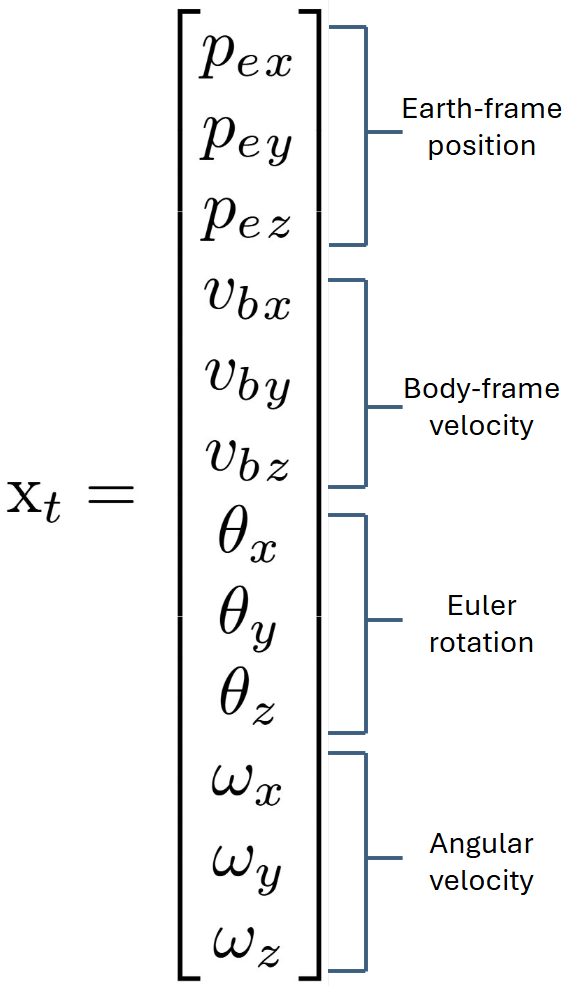}
    \caption{The \textbf{state representation} is designed to \textbf{capture all necessary dynamics of the system}, ensuring that it is sufficiently expressive for state prediction. This system is Markovian---given this state, external forces and moments can be inferred, eliminating hidden dependencies.}
    \label{fig:enter-label}
\end{SCfigure}

For the baseline world model, we implemented an implicit dynamics model consisting of an MLP initializer followed by an LSTM, as proposed in \citet{mohajerin_multi-step_2018}. This architecture takes as input a history of states and actions to initialize the hidden state and predicts the next state in the sequence. The model was trained using mean squared error (MSE) loss over predicted vs. ground-truth states.

\begin{figure}[h]
    \centering
    \includegraphics[width=1.0\linewidth]{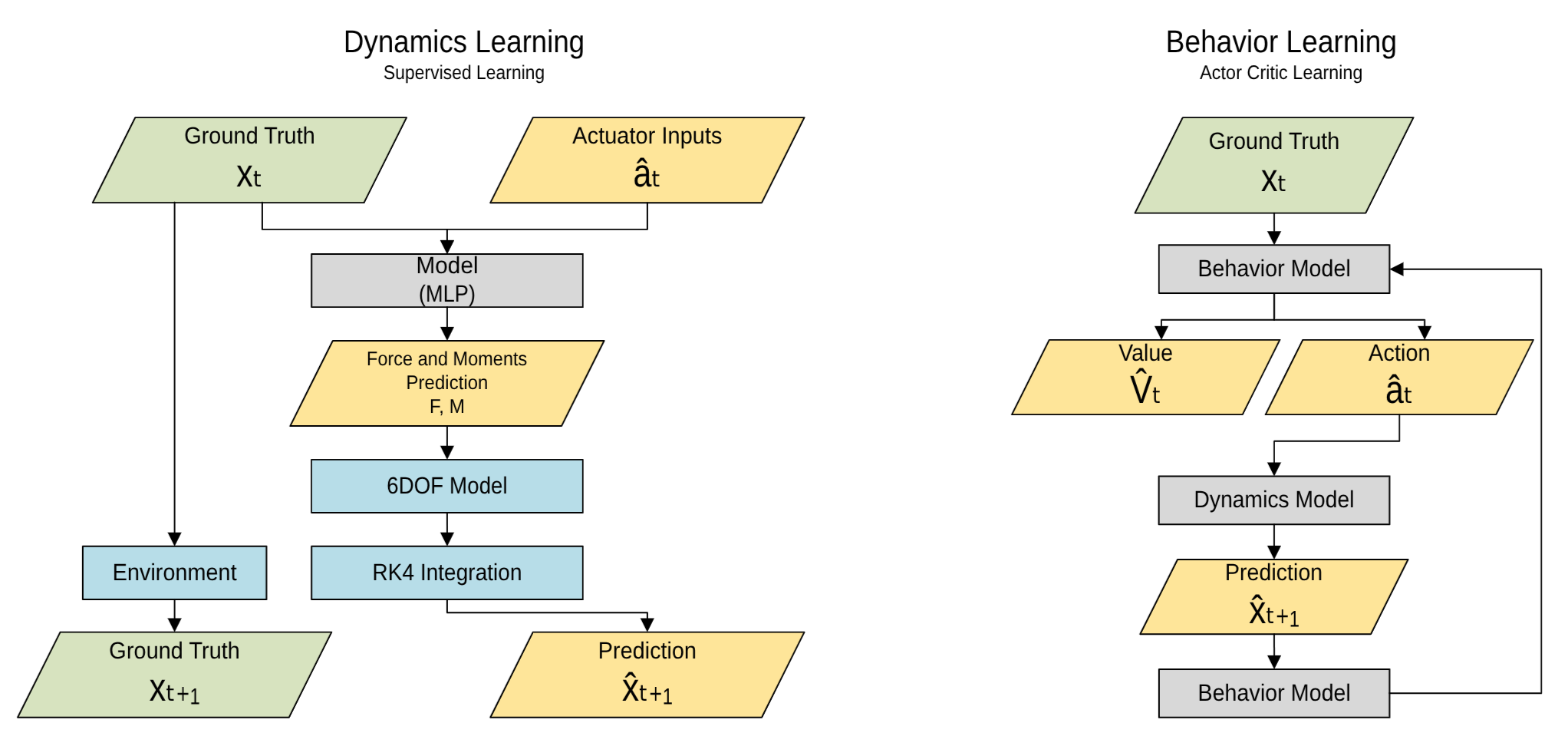}
    \caption{\textbf{Dreaming Falcon Algorithm Flowchart.} Our key contribution is that rather than directly predicting the next state, we use an MLP to estimate the net forces and moments acting on the free body---which are integrated to get the state.}
    \label{fig:alg-flowchart}
\end{figure}

Figure \ref{fig:alg-flowchart} illustrates our approach. In order to obtain the next state prediction, the MLP output (forces and moments) is passed through the 6DOF equations of motion \citep{matlab6dof} to compute the state derivative. We then apply the RK4 integration to obtain the next predicted state. The MSE loss is calculated between the predicted and ground truth states, and gradients are backpropagated through both the RK4 integrator and 6DOF dynamics to update the MLP parameters via gradient descent.

\begin{figure}[h]
    \centering
    \includegraphics[width=0.7\linewidth]{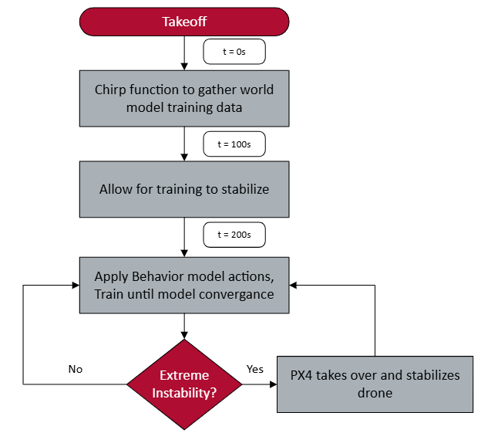}
    \caption{\textbf{Training process diagram.} We apply a chirp function with a fixed velocity trim for 100 seconds to collect initial data to load the replay buffer. Then, we train our world model and behavior model for 100 seconds to allow the training to stabilize. Finally, we deploy learned policies on the drone and update our behavior model in the loop.}
    \label{fig:training-pipeline}
\end{figure}

For training Dreaming Falcon, we used the training process shown in Figure \ref{fig:training-pipeline}. A key contribution of this training process is the use of a chirp function with a velocity trim to load the replay buffer.

To isolate the world model training, we applied a chirp function on the rate setpoint for 2.5 seconds with a randomized velocity trim. Then, we reset the drone to its origin using a position controller. We repeated this process for 100 seconds to load the replay buffer. Finally, we started training the world model for the next 10 minutes while repeating the chirps.
        
To parallel train the world model and behavior model, we ran the same chirp for the first 100 seconds to load the replay buffer. However, this time, we started applying behavior model actions and using actor-critic learning with straight through gradients to perform model updates. We continued this training until the model converged. For safety purposes, if the drone became unstable, the PID would take over and we would keep a count of the number of interrupts.

\section{Experiments \& Results}

To evaluate generalization, we assess model performance on both in-distribution (ID) rollouts sampled from the training replay buffer and out-of-distribution (OOD) rollouts representing novel trajectories not seen during training (e.g., transitions from hover to forward flight).

In order to highlight the shortcoming of this algorithm, we focus our experiments and results on the isolation of the world model. To demonstrate the key issues, all the experiments were run in the Gazebo simulator.

\begin{table}[h!]
\centering
\begin{tabular}{lcccc}
\toprule
\textbf{Position (Earth) (m)} & \textbf{N} & \textbf{E} & \textbf{D} & \textbf{Overall} \\
\midrule
RMSE & 0.054 & 0.049 & 0.049 & 0.051 \\
\midrule
\textbf{Velocity (Body) (m/s)} & \textbf{F} & \textbf{R} & \textbf{D} & \textbf{Overall} \\
\midrule
RMSE & 0.068 & 0.051 & 0.070 & 0.064 \\
\midrule
\textbf{Attitude (rad)} & \textbf{x} & \textbf{y} & \textbf{z} & \textbf{Overall} \\
\midrule
RMSE & 0.098 & 0.103 & 0.253 & 0.168 \\
\midrule
\textbf{Angular Velocity (rad/s)} & \textbf{x} & \textbf{y} & \textbf{z} & \textbf{Overall} \\
\midrule
RMSE & 0.339 & 0.448 & 0.199 & 0.344 \\
\bottomrule
\end{tabular}
\caption{\textbf{RMSE for state prediction on in-distribution (ID) validation rollouts using RNN world model.} After the model converged, the RNN model was applied to a 128 step validation rollout based on a sequence of actions randomly selected from the replay buffer. The validation score is RMSE between the prediction and ground truth states.}
\label{table:rnn-rmse}
\end{table}

\begin{table}[h!]
\centering
\begin{tabular}{lcccc}
\toprule
\textbf{Position (Earth) (m)} & \textbf{N} & \textbf{E} & \textbf{D} & \textbf{Overall} \\
\midrule
RMSE & 0.235 & 0.237 & 0.164 & 0.215 \\
\midrule
\textbf{Velocity (Body) (m/s)} & \textbf{F} & \textbf{R} & \textbf{D} & \textbf{Overall} \\
\midrule
RMSE & 0.295 & 0.280 & 0.251 & 0.276 \\
\midrule
\textbf{Attitude (rad)} & \textbf{x} & \textbf{y} & \textbf{z} & \textbf{Overall} \\
\midrule
RMSE & 0.137 & 0.150 & 0.082 & 0.126 \\
\midrule
\textbf{Angular Velocity (rad/s)} & \textbf{x} & \textbf{y} & \textbf{z} & \textbf{Overall} \\
\midrule
RMSE & 0.282 & 0.330 & 0.108 & 0.226 \\
\bottomrule
\end{tabular}
\caption{\textbf{RMSE for state prediction on in-distribution (ID) validation rollouts using physics-informed model.} Similarly to table \ref{table:rnn-rmse}, the physics-informed world model was applied to a 128 step validation rollout based on a sequence of actions randomly selected from the replay buffer. The validation score is RMSE between the prediction and ground truth states.}
\label{table:dreamer-rmse}
\end{table}

Tables \ref{table:rnn-rmse} and \ref{table:dreamer-rmse} show validation RMSE over a 128 step rollout after 5000 gradient updates (approximately 10 minutes of training).

Then the same model is applied to a test sequence recorded while the quadcopter transitioned from hover to forward flight for 50 rollout steps at 20 hz.

\begin{figure}[h]
    \centering
    \includegraphics[width=1.0\linewidth]{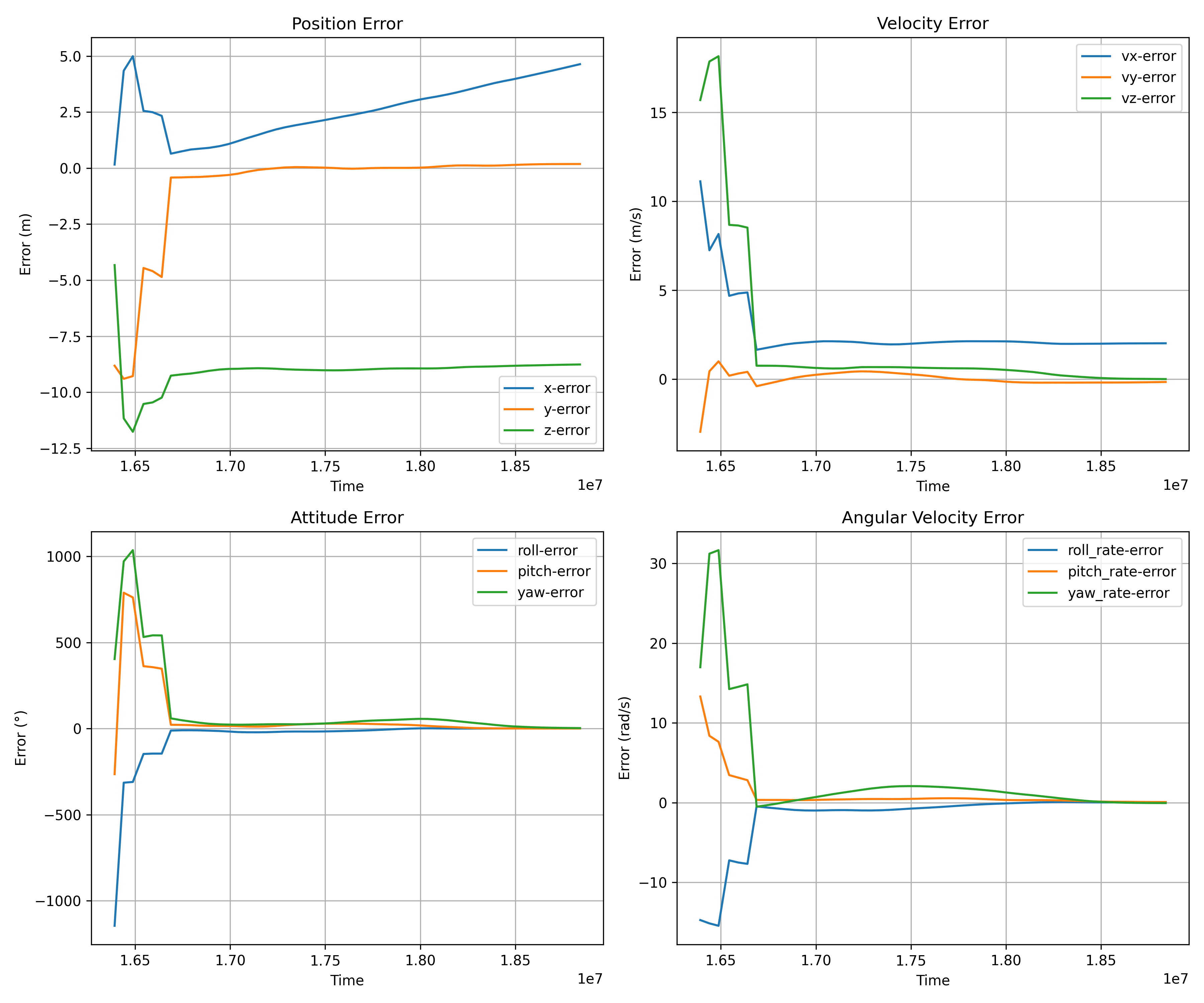}
    \caption{\textbf{RNN model state prediction error on out-of-distribution (OOD) test set: model rollout diverges during hover-to-forward transition.} We evaluate the RNN model performance using a 128 step rollout based on actions collected from the quadcopter switching from hover to forward flight. The errors between the ground truth and model predictions are illustrated in the figure.}
    \label{fig:RNNErrPlot}
\end{figure}

\begin{figure}[h]
    \centering
    \includegraphics[width=1.0\linewidth]{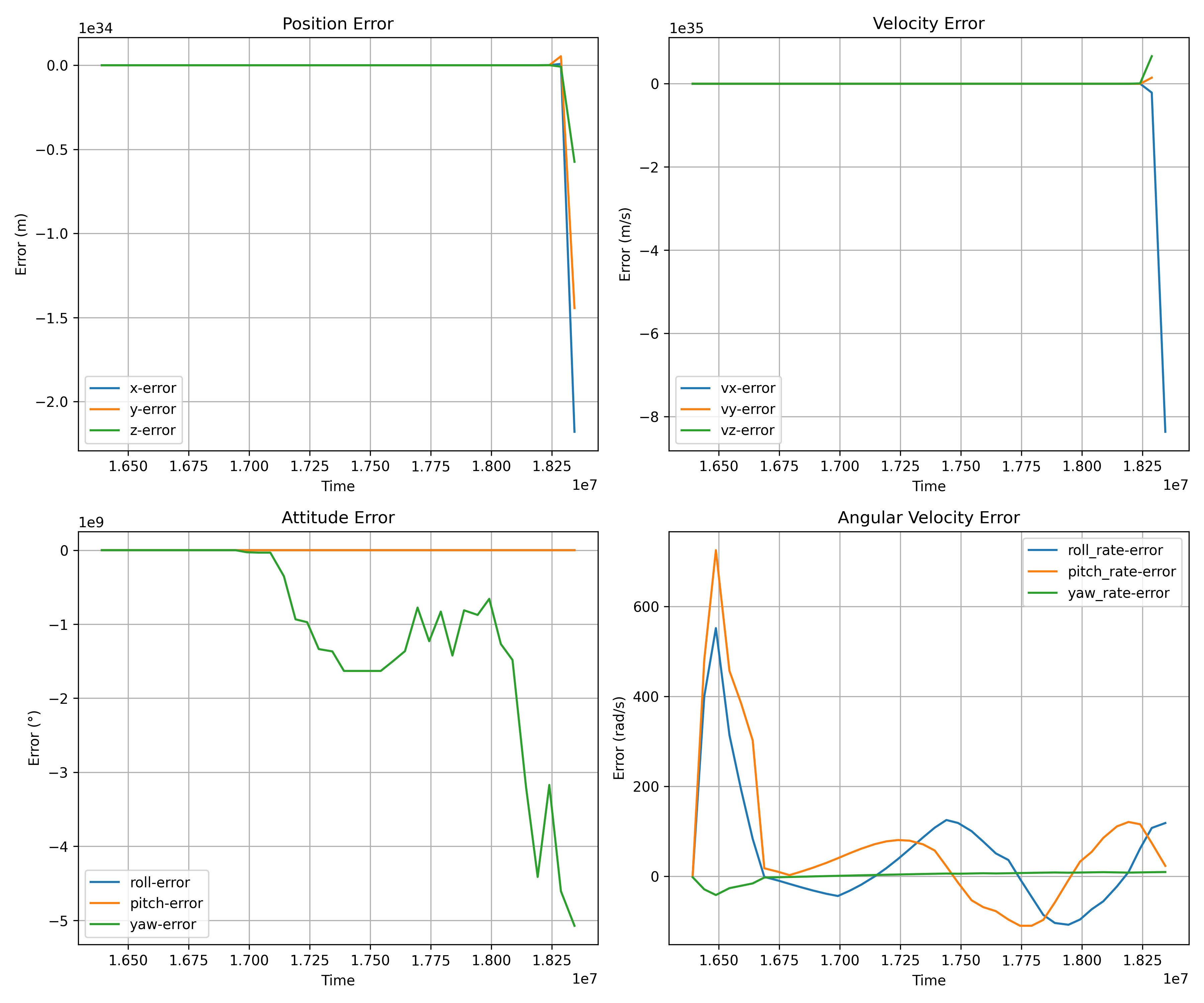}
    \caption{\textbf{Physics-informed model state prediction error on out-of-distribution (OOD) test set: model rollout diverges during hover-to-forward transition.} We evaluate the Physics Informed model performance using the same 128 step rollout as Figure \ref{fig:RNNErrPlot}. The errors between the ground truth and model predictions are illustrated in the figure.}
    \label{fig:DreamerErrPlot}
\end{figure}

Figure \ref{fig:DreamerErrPlot} shows errors of the state predictions on this test set plotted over time for both models.

While Tables \ref{table:rnn-rmse} and \ref{table:dreamer-rmse} show low RMSE on ID rollouts, Figures \ref{fig:RNNErrPlot} and \ref{fig:DreamerErrPlot} reveal that both models fail to generalize on OOD trajectories. Therefore, both the physics-informed and RNN based world models are able to accurately predict replay buffer data, however, fail to predict the test set accurately---which diverged over time. Due to this failure to generalize, it is reasonable to conclude that neither the physics-informed nor the RNN models are sample efficient enough to learn quadcopter dynamics.

When analyzing the issue further, we recognize that a majority of the data is in either hover, trim state, or chirp + trim state. The transition between these states is sparsely represented. Therefore, it is likely learning dynamics linearized around specific operating points, but failing to learn the transition between operating points---a critical component of modeling quadcopter dynamics. Our next step is to figure out an approach to augment the data to bias the training towards a more diverse set of data capturing all of the possible flight states and the transitions between them.

\section{Conclusion} 
\label{sec:conclusion}

Our hypothesis was that the Dreamer-style world model, when augmented with a physics-informed structure, would overcome the generalization challenges typically faced in quadcopter modeling. Specifically, we believed that predicting net forces and moments acting on the free body and integrating with a known 6DOF dynamics model would provide the inductive bias needed to stabilize world-model learning and enable robust policy training.

Unfortunately, our results show that this was not the case. Both the physics-informed model and the RNN-based world model performed well on the training data but did not generalize to new flight trajectories. This led to rapid divergence in rollout predictions, which prevented policy convergence. We hypothesize that the key issue is that the transition dynamics between different operating points of the vehicle is sparsely represented in the training data. Our next step is to investigate this hypothesis further.

These results highlight the ongoing difficulty of learning accurate world models for quadcopters---an example of an underactuated, high-dimensional system.

\section*{Acknowledgments}


\bibliographystyle{IEEEtranN}
\bibliography{references}

\end{document}